# Exponential scaling of neural algorithms – a future beyond Moore's Law?

James B Aimone
Center for Computing Research
Sandia National Laboratories
Albuquerque, NM USA
jbaimon@sandia.gov

*Abstract*— Although the brain has long been considered a potential inspiration for future computing, Moore's Law - the scaling property that has seen revolutions in technologies ranging from supercomputers to smart phones - has largely been driven by advances in materials science. As the ability to miniaturize transistors is coming to an end, there is increasing attention on new approaches to computation, including renewed enthusiasm around the potential of neural computation. This paper describes how recent advances in neurotechnologies, many of which have been aided by computing's rapid progression over recent decades, are now reigniting this opportunity to bring neural computation insights into broader computing applications. As we understand more about the brain, our ability to motivate new computing paradigms with continue to progress. These new approaches to computing, which we are already seeing in techniques such as deep learning and neuromorphic hardware, will themselves improve our ability to learn about the brain and accordingly can be projected to give rise to even further insights. This paper will describe how this positive feedback has the potential to change the complexion of how computing sciences and neurosciences interact, and suggests that the next form of exponential scaling in computing may emerge from our progressive understanding of the brain.

*Keywords—Moore's Law, neuromorphic computing, neuroscience, BRAIN Initiative, scaling*

## I. The End of Moore's Law and the Renewed Relevance of the Brain

The impending demise of Moore's Law has begun to send shockwaves through the computing research community[1]. Moore's Law has driven the computing industry for many decades, with nearly every aspect of society benefiting from the nearly tireless advance of improved computing processors, sensors, and controllers. Behind these products has been a considerable research industry, with billions of dollars invested in fields ranging from computer science to electrical engineering. Fundamentally, however, the exponential growth in computing described by Moore's Law was driven by advances in materials science[2, 3]. From the start, the power of the computer has been limited by the density of transistors. Progressive advances in how to manipulate silicon through advancing lithography methods and new design tools have kept advancing computing in spite of perceived limitations of the dominant fabrication processes of the time[3].

There is strong evidence that this time is indeed different, and Moore's Law is soon to be over for good[1, 4]. Already, Dennard scaling, Moore's Law's lesser known but equally important parallel, appears to have ended[5]. Dennard's scaling refers to the property that the reduction of transistor size (due to Moore's law) came with an equivalent reduction of required power[6]. This has real consequences – even though Moore's law has continued over the last decade, allowing feature sizes to go from ~65nm to less than 10nm; the ability to speed up processors for a constant power cost has stopped. Today's common CPUs are limited to about 4GHz due to heat generation, which is roughly the same as they were 10 years ago. While Moore's Law enables more CPU cores on a chip (and has enabled high power systems such as GPUs to continue advancing), there is increasing appreciation that feature sizes cannot fall much further, with perhaps 2 or 3 further generations remaining prior to ending.

Multiple solutions have been presented for technological extension of Moore's Law[1, 4, 7, 8], but there are two main challenges that need addressing. For the first time, it is not immediately evident that future materials will be capable of providing a long-term scaling future– while non-silicon approaches such as carbon nanotubes or superconductivity may yield some benefits, these approaches also face theoretical limits that are only slightly better than the limits CMOS is facing[9]. Somewhat more controversial, however, is the observation that requirements for computing are changing[7, 8]. In some respects, the current limits facing computing lie beyond what the typical consumer outside of the high-performance computing community ever will require for floating point math. Data-centric computations such as graph analytics, machine learning, and searching large databases are increasingly pushing the bounds of our systems and are more relevant for a computing industry built around mobile devices and the Internet. As a result, it is reasonable to consider that the ideal computer is not one that is better at more FLOPS, but rather one that is capable of providing low-power computation more appropriate for a world flush with "big data".

For these reasons, neural computing has begun to gain increased attention as a post-Moore's law technology. In many respects, neural computing is an unusual candidate to help extend Moore's Law. Neural computing is effectively an

The authors acknowledge financial support from the DOE Advanced Simulation and Computing program and Sandia National Laboratories' Laboratory Directed Research and Development Program. Sandia National Laboratories is a multimission laboratory managed and operated by National Technology and Engineering Solutions of Sandia, LLC., a wholly owned subsidiary of Honeywell International, Inc., for the U.S. Department of Energy's National Nuclear Security Administration under contract DE-NA0003525.

XXX-X-XXXX-XXXX-X/XX/$XX.00 ©20XX IEEE

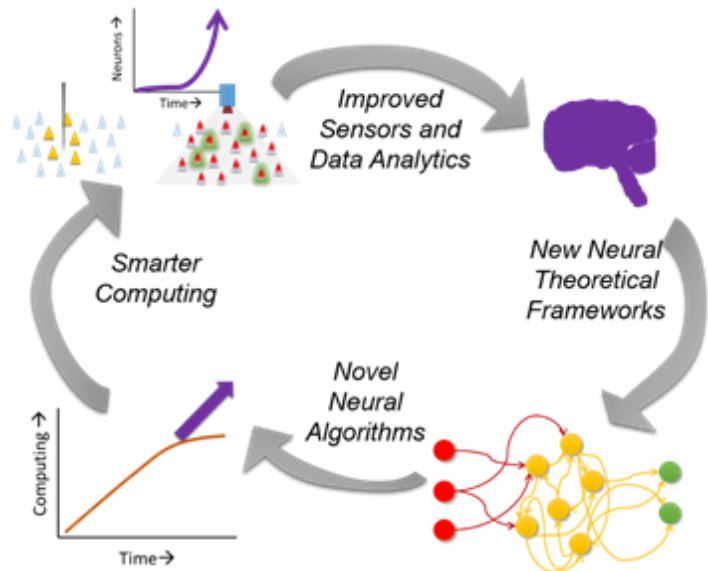

**Fig. 1**. Moore's Law has helped initiate a potential positive feedback loop between neural data collection and improved computation. Moore's Law has enabled the miniaturization of sensors and improved analytics necessary to improve neural data collection. This burst of neural data has the potential to dramatically improve our ability to extract knowledge from the brain. The improved neural insights can motivate improved neural computation algorithms and architectures; ant these in turn can help further the advances of computing.

algorithmic and architectural change from classic numerical algorithms on von Neumann architectures, as opposed to exploiting a novel material to supplant silicon. Further, unlike quantum computation, which leverages different physics to perform computation, neural computing likely falls within the bounds of classic computing theoretical frameworks. Whereas quantum computation can point to exponential benefits on certain tasks such as Shor's quantum algorithm for factoring numbers[10]; neural computing architecture's most likely path to impact is through polynomial tradeoffs between energy, space, and time. Such benefits can be explicitly formalized and are potentially quite impactful for certain applications[11], but neural architectures are unlikely going to be fundamentally more powerful on generic applications than the general purpose architectures used today[8].

It is worth asking whether neural computing is truly a paradigm that will permit exponential scaling going forward, or rather would it represent a "one-off" gain of efficiency in some dimension, such as power efficiency? While potentially impactful, such a value proposition would not represent a long-lasting scaling capability. While to some the distinction between these two futures may appear semantic, there is a considerable difference. If neural architectures indeed represent only a one-time gain to accelerate a handful of algorithms, then it perhaps merits some consideration by specialized communities; however, if neural computation actually can represent a scalable technology, it would permit the trillion-dollar computing industry to redirect its considerable resources towards it.

This article posits that neural-inspired computation represents a distinctly new approach to scaling computing technology going forward. Instead of relying on continual advances in miniaturization of devices; neural computing is positioned to benefit from long-lasting intellectual advances due to our parallel gain of knowledge of the brain's function (**Fig. 1**). In effect, because the materials science and chemistry of devices has been so extensively optimized, we are better positioned to mine the brain for neural inspiration and hopefully achieve a continual advancement of our neural computing capabilities through algorithmic and architectural advances.

## II. KNOWLEDGE OF THE BRAIN IS UNDERGOING ITS OWN DRAMATIC SCALING

Several critical efforts are underway that make such a revolutionary perspective possible. Major efforts in neuroscience, such as the BRAIN Initiative in the United States and the Human Brain Project in the European Union, are focused directly on extracting the type of neuroscience knowledge that will be useful for computing[12-14]. Specifically, the BRAIN Initiative has a guiding goal of recording a million neurons simultaneously in awake, behaving animals[14]. Such a goal would have been unfathomable only a few years ago; however today it increasingly appears within neuroscientists' grasp. Somewhat ironically, the advances in neuroscience sensors that allow neuroscientists to measure the activity of thousands of neurons at a time have been fueled in large part by the miniaturization of devices due to Moore's law. Indeed, it has been noted that the increase in numbers of neurons recorded within a single experiment has itself undergone an exponential scaling over recent decades[15].

Similarly, large scale efforts seeking to reconstruct the "connectome" of the brain are becoming more common[16]. In contrast to artificial neural networks (ANNs), neural circuits are highly complex and vary considerably across brain regions and across organisms. This connectome effectively represents the graph on which biological neural computation occurs, and understanding this connectivity is arguably critical for understanding the wide range of neural computations that are available for computing inspiration. While the technology to image these large scale connectomes is increasingly within neuroscientist's grasp, there is a growing appreciation that challenges surrounding data analysis and storage are likely to become the limiting factor of neurotechnology as opposed to simply achieving higher resolution sensor technologies [17-19].

Alongside this robust growth of neuroscience data, and a corresponding increase in knowledge of the brain, has been an explosion in the amount of raw data in the world. Fueled in large part by the Internet, the amount of data available to train machine learning tools such as deep networks is no longer one of the primary limitations to algorithm design[20]. Just like the growth in neural data, this increase in broader data about the world has occurred in large part due to Moore's Law and its dependent technology developments. The combination of large data sets and computing power allowed classic ANNs to achieve their long-touted promise. It is possible, if not likely, that the combination of increased neural knowledge with sufficient quantities of data could be equally instrumental in enabling the development of other novel neural-inspired algorithms. Indeed, new algorithms inspired by increased neural knowledge of regions not classically associated with machine learning, like the hippocampus, pre-frontal cortex, or striatum, may well represent a disruptive advance over both existing conventional computing and data-centric computing capabilities.

### III. WHAT WOULD EXPONENTIAL SCALING OF NEURAL ALGORITHMS LOOK LIKE?

#### A. Is data efficiency a potential metric for neural scaling?

The primary challenge to this perspective is one of quantification – *what does it mean for a brain-inspired neural algorithm's performance to scale?* When Gordon Moore first presented his scaling law, there was already empirical evidence suggesting that materials would continue scaling, and there were clear metrics for illustrating this development[3]. Nothing comparable to these device-level metrics exist for neural inspiration - even comparing algorithm to algorithm presents a challenge – for instance what basis can be used to compare a loosely visual cortex-inspired deep learning neural network algorithm to an auditory nuclei-inspired dynamic sound localization algorithm? Nor do conventional computing metrics, such as floating point operations (FLOPS), make a lot of sense when discussing higher level algorithmic performance.

Ideally, a powerful metric for describing the advancement of neural algorithms will both be quantifiable and general. If we focus exclusively on the classes of neural algorithms being implemented today, one potential metric for ascertaining the relative value of future algorithms could be the required volume of high-quality training data. A notable example of this is deep learning. In only a few years, deep learning networks have become state of the art for many static sensory processing applications[20], however the success of deep learning is in large part attributable to the presence of high quality and often very large image data sets such as MNIST, CFAR, and ImageNet. While large volumes of data are increasingly common in many applications, obtaining high-quality data – defined by both well-calibrated sensors and effective annotations – is often incredibly expensive and time consuming. As a result, the ability for deep learning-related methods to impact domains with inappropriately structured data has been limited. Some of these goals are areas of intensive machine learning research today, but there are many reasons to believe that the brain's approach to maximizing the utility of observed data in both developmental and adult learning is a notable area where brain-inspiration can dramatically improve computing. Thus, one can envision the efficacy of future algorithms scaling with respect to data as opposed to power.

Although a neural scaling law centered around data efficiency would be attractive since it is quantifiable and concrete, it is unclear whether this will generalize beyond the current generation of brain-inspired algorithms. Ultimately, the power of Moore's Law as a description of computing scaling arose from the it being somewhat agnostic to the underlying materials that drove the technology advances. If a given material or design process saturated its utility, an alternative could just as readily step in and provide its own local contribution to the broader community. Similarly, any perspective on neural technology advancement should be robust to the inevitable saturation of a given algorithmic approach. If and when standard deep learning networks begin to experience only marginal advancements, which arguably is already occurring with increased attention to time-dependent networks, we should be ready to look to the next technology for its contribution.

#### B. Future scaling of neural algorithms will likely provide new capabilities as opposed to improving existing capabilities

Thus, the challenge to extrapolating the potential of neural computation to the future rests on finding a metric that can both be quantified and is robust to not only what neural algorithms do today, but what brain-inspired algorithms will be in the future. For conventional computing, the FLOP is a relatively ideal metric. It is simple enough to be readily mapped to conventional architectures, with clear relationships to underlying gate counts and transistor densities, while powerful enough to be relevant to almost all classic computational functions. Beyond that, however, FLOPS are a simplistic description of what computers are often used for: math.

For neural algorithms, and thus the associated hardware we potentially will use to accelerate them, such a metric really should quantify something akin to intelligence. For this reason, it is unsatisfying to simply quantify neuronal or synaptic operation efficiency and call it a day. While these metrics take inspiration from the low-level structure of the brain, it is not clear how they directly map to even today's neural algorithms, much less those in the future. In this sense, they are more akin to gates or transistor counts than FLOPS; with the caveat that because we do not currently know the neural equivalent of a FLOP, we cannot say for certain how these low level neural gates will relate to future neural algorithm functions.

The following sections describe a potential future outlook for neural scaling from the other direction – the development of progressively more advanced brain-inspired capabilities in algorithms. These sections come from the perspective that the rapid increase in available experimental data is a trend that is unlikely to end soon. While the BRAIN Initiative goal of simultaneously recording a million neurons sounds impressive, it is worth remembering that is only a tiny fraction of a rodent cortex; and the diversity of neural regions and complex behaviors suggests that a plethora of algorithms wait to be defined. A major indicator of this progress will be potential developments in theoretical neuroscience. Neuroscience has long been a field where the ability to collect data has constrained the development of robust neural theories, and it is a growing hope that the ability to measure large populations of neurons simultaneously will inspire the development of more advanced neural theories that that previously would have been dismissed as non-falsifiable due to our inability to perform the requisite experiments in the brain[21-23].

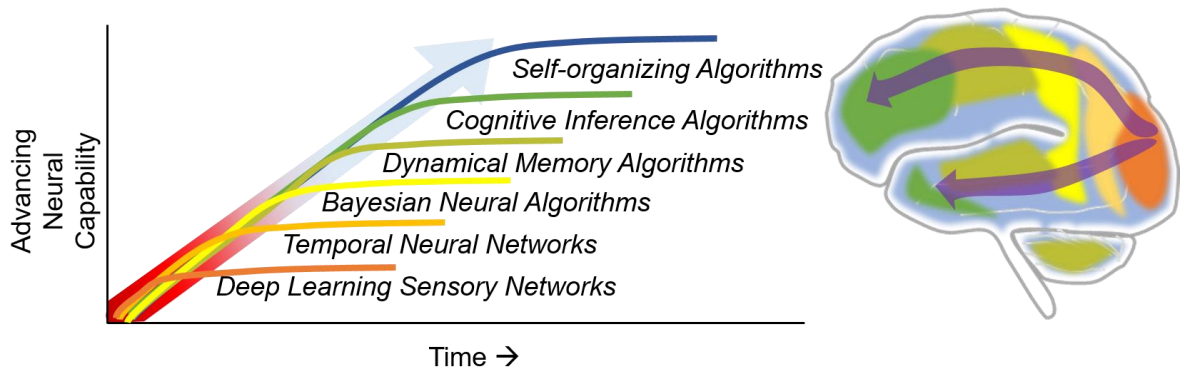

| Algorithm Class | Example Algorithms | Inspiration | Application |
| --- | --- | --- | --- |
| Deep Learning | Deep Convolutional Networks (VGG, AlexNet, GoogleNet, etc.), HMax, Neocognitron | Hierarchy of sensory nuclei and early sensory cortices | Static feature extraction (e.g., images) & pattern classification |
| Temporal Neural Networks | Deep Recurrent Networks (long short-term memory), Hopfield Networks | Local recurrence of most biological neural circuits, especially higher sensory cortices | Dynamic feature extraction (e.g., videos, audio) & classification |
| Bayesian Neural Algorithms | Predictive Coding, Hierarchical Temporal Memory | Substantial reciprocal feedback between "higher" and "lower" sensory cortices | Inference across spatial and temporal scales |
| Dynamical Memory and Control Algorithms | Liquid State Machines, Echo State Networks, Neural Engineering Framework | Continual dynamics of hippocampus, cerebellum, and prefrontal and motor cortices | Online learning content-addressable memory & adaptive motor control |
| Cognitive Inference Algorithms | Deep reinforcement learning (Q-learning) | Integration of multiple modalities and memory into prefrontal cortex, which provides top-down influence on sensory processing | Context and experience dependent information processing and decision making |
| Self-organizing Algorithms | Neurogenesis Deep Learning | Initial development and continuous refinement of neural circuits to specific input and outputs | Automated neural algorithm development for unknown input and output transformations |

**Fig. 2.** The continued scaling of neural computing need not rely on improved materials, but rather is based on the ongoing improvements of algorithmic potential. Today, we are exploiting advances of conventional artificial neural networks at large scale, but there are already trends towards more temporal based neural networks such as long short-term memory. We are poised to benefit from a series of these technological advances, brining neural algorithms closer to the more sophisticated computational potential of the brain.

**Fig. 2** illustrates one potential path by which neural algorithms could scale going forward. Because of the aforementioned advances in neuroscience, it is reasonable to expect that the current deep learning revolution could be followed by complementary revolutions in other cognitive domains. Each of these novel capabilities will build on its predecessors – it is unlikely that any of these algorithms will ever go away, but it will continue to be used as an input substrate for more sophisticated cognitive functions.

*1) Deep Learning*

Deep learning, as defined here, is not a new technology. While there have been a few key theoretical advances, at their core deep learning networks, such as deep convolutional networks, are not fundamentally different from ANNs being developed in the 1980s and 1990s. As mentioned before, the success of deep networks in the past decade has been driven in large part due to the availability of sufficiently rich data sets as well as the recognition that modern computing technology, such as GPUs, are effective at training at large scale. In many ways, these advances that have enabled deep learning have simply allowed ANNs to realize the potential that cognitive scientists have been espousing for half a century.

From a neuroscience perspective, deep learning's success is both promising and limited. The pattern classification function that deep networks excel at is only a very narrow example of cognitive functionality, albeit one that is quite important. The inspiration it takes from the brain is quite restricted as well. Deep networks are arguably inspired by neuroscience that dates to the 1950s and 1960s, with the recognition by neuroscientists like Vernon Mountcastle, David Hubel and Torsten Wiesel that early sensory cortex is modular, hierarchical, and has representations that start simple in early layers and become progressively more complex. While these are critical findings that have also helped frame cortical research for decades, the neuroscience community has built on these findings in many ways that have yet to be integrated into machine learning. One such example, described next, is the importance of time.

*2) Temporal Neural Networks*

Already we appear to be at a transition point in the neural algorithm community. Today, much of the research around neural algorithms is focused on extending methods derived from deep learning to operate with temporal components. These methods, including techniques such as long short-term memory, are quickly beginning to surpass state of the art on more time-dependent tasks such as audio processing[20]. Similar to more conventional deep networks, many of these time-based methods leverage relatively old ideas in the ANN community around using network recurrence and local feedback to represent time.

While this use of time arising from local feedback is already proving powerful, it is a fairly limited implementation of the temporal complexity within the brain. Local circuits are incredibly complex; often numerically dominating inputs and outputs to a region and consisting of many distinct neuron types [24]. The value of this local complexity likely goes far beyond the current recurrent ANN goals of maintaining a local state for some period of time. In addition to the richness of local biological complexity, there is the consideration of what spike based information processing means with regard to contributing information about time. While there is significant discussion around spike-based neural algorithms from the perspective of energy efficiency; less frequently noted is the ability of spiking neurons to incorporate information in the time domain. This use of time can be very explicit, as we have shown in neural algorithms that use spike timing to efficiently represent information for algorithms such as comparing cross-correlations [25].

Extracting more computational capabilities from spiking and the local circuit complexity seen in cortex and other regions has the potential to enable temporal neural networks to continue to become more powerful and effective in the coming years. However, it is likely that the full potential of temporal neural networks will not be fully realized until they are fully integrated into systems that also include the complexity of regional communication in the brain, such as networks configured to perform both top-down and bottom-up processing simultaneously, such as neural-inspired Bayesian inference networks.

*3) Bayesian Neural Algorithms*

Even perhaps more than the time, the most common critique from neuroscientists about the neural plausibility of deep learning networks is the general lack of "top-down" projections within these algorithms. Aside from the optic nerve projection from retina to the LGN area of the thalamus, the classic visual processing circuit of the brain includes as much, and often more, top-down connectivity between regions (e.g., V2→V1) as it contains bottom-up (e.g., V1→V2).

Not surprisingly, the observation that higher-level information can influence how lower-level regions process information has strong ties to well established motifs of data processing based around Bayesian inference. Loosely speaking, these models allow data to be interpreted not simply by low-level information assembling into higher features unidirectionally, but also by what is expected – either acutely based on context or historically based on past experiences. In effect, these high-level "priors" can bias low-level processing towards more accurate interpretations of what the input means in a broader sense.

While the extent to which the brain is perfectly explained by this Bayesian perspective is continually debated, it is quite clear that the brain does use higher-level information, whether from memory, context, or cross-sensory modalities, to guide perception of any sensory modality. If you expect to see a cloud shaped like a dog, you are more likely to see one. The application of these concepts to machine learning has been more limited, however. There are clear cases of non-neural computer vision algorithms based on Bayesian inference principles[26], though it has been challenging to develop such models that can be trained as easily as deep learning networks. Alternatively, other algorithms, such as Hierarchical Temporal Memory (HTM)[27] and predictive networks (PredNet)[28] have been developed that also leverage these top-down inputs to drive network function. These approaches are not necessarily explicitly Bayesian in all aspects, but do indicate that advances in this area are occurring.

Ultimately, however, this area will be enabled by increased knowledge about how different brain areas interact with one another. This has long been a challenge to neuroscientists, as most experimental physiology work was relatively local and anatomical tracing of connectivity has historically been sparse. This is changing as more sophisticated physiology and connectomics techniques are developed. For example, the recently proposed technique to "bar-code" neurons uniquely could enable the acquisition of more complete, global graphs of the brain [29].

Of course, the concept of Bayesian information processing of sensory inputs, like the previous two algorithmic frameworks described before, is skewed heavily towards conventional machine learning tasks like classification. However, as our knowledge of the brain becomes more extensive, we can begin to take algorithmic inspiration from beyond just sensory systems. Most notable will be dynamics and memory.

*4) Dynamical memory and control algorithms*

As the previous sections have alluded to, biological neural circuits have both greater temporal and architectural complexity than classic ANNs. Beyond just being based on spikes and having feedback, it is important to consider that biological neurons are not easily modeled as discrete objects like transistors, rather they are fully dynamical systems exhibiting complex behavior over many state variables. While considering biological neural circuits as complex assemblies of many dynamical neurons whose interactions themselves exhibit complex dynamics seems intractable as an inspiration for computing, it is worth noting that there is increasing evidence that it is possible to extract computational primitives from such neural frameworks, particularly when anatomy constraints are considered. Increasingly, algorithms like liquid state machines (LSMs)[30] and echo-state networks[31] have been introduced that abstractly emulate cortical dynamics loosely by balancing activity in neural circuits that exhibit chaotic (or near chaotic) activity. Alternatively, by appreciating neural circuits as programmable dynamical systems, approaches like the Neural Engineering Framework (NEF) have shown that complex dynamical algorithms can be programmed to perform complex functions[32].

While these algorithms have shown that dynamics can have a place in neural computation, the real impact from the brain has yet to be appreciated. Neuroscientists increasingly see regions like the motor cortex, cerebellum, and hippocampus as being fundamentally dynamical in nature: it is less important what any particular neuron's average firing rate is, and more important what the trajectory of the population's activity is.

The hippocampus makes a particularly interesting case to consider here. Early models of the hippocampus were very similar to Hopfield networks – memories were represented as auto-associative attractors that could reconstruct memories from a partial input. These ideas were consistent with early place cell studies, wherein hippocampal neurons would fire in specific locations and nowhere else. While a simple idea to describe, it is notable how for roughly forty years this idea has failed to inspire any computational capabilities. However, increasingly it is appreciated that the hippocampus is best considered from a dynamical view: place cell behavior has long been known to be temporally modulated and increasing characterization of "time cells" is indicative that a more dynamical view of hippocampal memory is likely a better description of hippocampal function and also potentially more amenable to inspiring new algorithms.

Of course, developing neural-inspired dynamical memory and control algorithms has the potential to greatly advance these existing techniques, but the real long-lasting benefit from neural computing will likely arise when neuroscience provides the capability to achieve higher-level cognition in algorithms.

*5) The unknown future: cognitive inference algorithms & self-organizing algorithms & beyond*

Not coincidentally, the description of these algorithms has been progressing from the back of the brain towards the front, with an initial emphasis on early sensory cortices and eventually progressing to higher level regions like motor cortex and the hippocampus. While neural machine learning is taking this back-to-front trajectory, these areas have all received reasonably strong levels of neuroscience attention historically – the hippocampus arguably is as well studied as any cortical region. The "front" of the brain, in contrast, has continually been a significant challenge to neuroscientists. Areas such as the prefrontal cortex and its affiliated subcortical structures like the striatum have remained a significant challenge from a systems neuroscience level, in large part due to their distance from the sensory periphery. As a result, behavioral studies of cognitive functions such as decision making are typically highly controlled to eliminate any early cortical considerations. Much of what we know from these regions originates from clinical neuroscience studies, particularly with insights from patients with localized lesions and neurological disorders such as Huntington's and Parkinson's diseases.

As a result, it is hard to envision what algorithms inspired by prefrontal cortex will look like. One potential direction are recent algorithms based on deep reinforcement learning, such as AlphaGo's deep Q-learning, which has been successful in winning against humans in games of presumably complex decision-making[33]. These deep reinforcement algorithms today are very conventional, only touching loosely on the complexity of neuromodulatory systems such as dopamine that are involved in reinforcement learning, yet they are already disrupting all of the long-held challenges in artificial intelligence.

While learning more about the frontal and subcortical parts of the brain offer the potential to achieve dramatic capabilities associated with human cognition, there is an additional benefit that we may achieve from considering the longer time-scales of neural function, particularly around learning. Most neural algorithms learn through processes based on synaptic plasticity. Instead, we should consider that the brain learns at pretty much all relevant time-scales – over years even in the case of hippocampal neurogenesis and developmental plasticity. Truly understanding how this plasticity relates to computation is critical for ever fully realizing the true potential of neural computation.

We and others have begun to scrape the surface of what lifelong learning in algorithms could offer with early signs of success. For instance, adding neurogenesis to deep learning enables these algorithms to function even if the underlying data statistics change dramatically over time[34]. The recently announced Lifelong Learning in Machines program by DARPA promises to explore this area deeper as well.

*C. Progress in neural algorithms will build on itself*

This progression of neural functionality very well could make combined systems considerably more powerful than the individual components. If a hippocampal-inspired one-shot learning algorithm could enable the continuous adaptation of a deep learning network, it could considerably increase the long-term utility of that algorithm [34]. This consideration of amortized cost of an algorithm is of course a very different approach to evaluating the costs and benefits of computing, but the prospect of continuously learning neural systems will require some sort of long-term evaluation. Further, just as the brain uses the hippocampus to provide a short-term memory function to complement the long-term memory of several sensory cortices, it is likely that future neural systems could be constructed in a modular manner whereby different combinations of neural components can amplify the performance on different functions.

While this discussion has focused primarily on the long-term benefits of modular neural algorithms, this predicted succession of algorithmic capabilities would be well positioned to be amplified by corresponding advances in computing architectures and materials[35]. Today, deep learning has already begun to substantially influence the design of computer architectures such as GPUs and implicitly underlying devices and materials. While materials have often been researched with respect to the ubiquitous binary transistor function common to von Neumann architectures, new architectures inspired by novel neural algorithm classes may introduce entirely new desirable characteristics for devices. For example, dynamical neural algorithms inspired by prefrontal and motor cortex may be best implemented on more dynamics-friendly devices capable of smooth state transitions as opposed to the very stiff and reliable operational characteristics of transistors today. One particular area where neural architectures could begin to have dramatic impact would be intrinsic capabilities for learning and self-organization[36]. While we are still a long-way away from understanding neural development from a computational theory perspective, the availability of such functionality at an architectural level will likely be very disruptive, particularly as algorithms leveraging more brain-like plasticity mechanisms are introduced [34].

## IV. CAN NEUROSCIENCE REALLY DRIVE COMPUTING LONG-TERM? WHAT HAS TO BE DONE?

While the above paragraphs lay out an argument for why neural computing could provide the computing industry with a future beyond Moore's Law, by no means is this future assured. Aside from the clear technical challenges that lie ahead related to implementing the intellectual trajectory laid out above; there are considerable social challenges that must be addressed as well.

Arguably, the greatest urgency is to inspire the broader neuroscience community to pursue developing theories that can impact neural computing. While there is considerable reason to believe that our knowledge of the brain will continue to accelerate through improved neurotechnologies, the path by which that knowledge can be leveraged into a real impact on computing is not well established. In particular, it is notable that much of the deep learning revolution underway was driven by computer scientists and cognitive scientists basing algorithms primarily on concepts well established in neuroscience in the 1940s and 1950s [37]. There are several examples to have optimism, however. The recently kicked off MICrONS program, which is part of the US BRAIN Initiative, aims directly at the challenge of leveraging high-throughput neuroscience data in novel algorithm development [38]. Google's DeepMind – a company started by cognitive neuroscientists – is at the forefront of successfully integrating neural concepts such as reinforcement learning into machine learning algorithms [33]. While the levels of neuroscience buy-in have been mixed, the EU Human Brain Project has been successful at renewing interest in neuromorphic technologies in the computer science and electrical engineering communities.

Nevertheless, there needs to be a more robust investment by neuroscientists if computing is to benefit from the revolutions underway in experimental neuroscience. This is particularly important if neural influence is to move beyond computer vision – a community which has had historic ties to their neuroscience vision researchers. For example, despite a historic level of attention and understanding that is roughly comparable to that of visual cortex [39], the hippocampus has had arguably very little influence on computing technologies, with only limited exploration of hippocampal-inspired spatial processing in SLAM (Simultaneous Localization and Mapping) applications [40] and almost no influence on computer memory research.

A renewed focus by neuroscientists on bringing true brain-inspiration to computation would be consistent with the field's broader goals in addressing the considerable mental health and neurological disorders facing society today [14]. Many of the clinical conditions that drive neuroscience research today can be viewed as impairments in the brain's internal computations, and it is not unreasonable to argue that taking a computing-centric perspective to understanding neurologically critical brain regions such as the striatum and hippocampus could facilitate new perspectives for more clinically-focused research [41].

A second, related challenge is the willingness of the computing communities to incorporate inspiration from a new source. For decades, computing advances have been driven by materials, with reduced emphasis on addressing the underlying von Neumann architecture. Given the perceived plateauing of this classic path, there is now considerable investment in neural architectures; efforts such as IBM TrueNorth [42] and the SpiNNaker [43] and BrainScales [44] systems out of the EU HBP have focused on powerful architectural alternatives in anticipation of neural algorithms. Other more-device driven efforts are focused on using technologies such as memristors to emulate synapses [36]. To some extent, these approaches are seeking to create general purpose neural systems in anticipation of eventual algorithm use; but these approaches have had mixed receptions due to their lack of clear applications. It is reasonable to expect that new generations of neural algorithms can drive these architectures going forward, but the parallel development of new strategies for neural algorithms with new architecture paradigms is a continual challenge.

One implication of the general disconnect between these very different fields is that few researchers are sufficiently well versed across all of these critical disciplines to avoid the at times detrimental misinterpretation of knowledge and uncertainty from one field to another. Questions such as "Are spikes necessary?" have quite different meanings to a theoretical neuroscientist and a deep learning developer. Similarly, few neuroscientists consider the energy implications of complex ionic Hodgkin-Huxley dynamics of action potentials, however many neuromorphic computing studies have leveraged them in their pursuit of energy efficient computing. Ultimately, these mismatches demand that new strategies for bringing 21$^{st}$ century neuroscience expertise into computing be explored. New generations of scientists trained in interdisciplinary programs such as machine learning and computational neuroscience may offer a long-term solution, but in the interim it is critical that researchers on all sides are open to the considerable progress made these complex, well established domains in which they are not trained.

ACKNOWLEDGMENT

The author thanks Dr. Kris Carlson and Dr. Erik Debenedictis for critical comments and discussions.

REFERENCES

[1] M. M. Waldrop, "The chips are down for Moore's law," *Nature News,* vol. 530, p. 144, 2016.
[2] G. E. Moore, "Progress in digital integrated electronics," in *Electron Devices Meeting*, 1975, pp. 11-13.
[3] S. E. Thompson and S. Parthasarathy, "Moore's law: the future of Si microelectronics," *Materials today,* vol. 9, pp. 20-25, 2006.

[4] S. I. Association and S. R. Corporation, "Rebooting the IT Revolution: A Call to Action," 2015.

[5] H. Esmaeilzadeh, E. Blem, R. S. Amant, K. Sankaralingam, and D. Burger, "Dark silicon and the end of multicore scaling," in *Computer Architecture (ISCA), 2011 38th Annual International Symposium on*, 2011, pp. 365-376.

[6] R. H. Dennard, F. H. Gaensslen, V. L. Rideout, E. Bassous, and A. R. LeBlanc, "Design of ion-implanted MOSFET's with very small physical dimensions," *IEEE Journal of Solid-State Circuits,* vol. 9, pp. 256-268, 1974.

[7] R. S. Williams, E. P. DeBenedictis, I. Arvind Kumar, M. Stalzer, M. Badaroglu, G. W. B. Qualcomm*, et al.*, "OSTP Nanotechnology-Inspired Grand Challenge: Sensible Machines (extended version 2.5)," 2015.

[8] J. M. Shalf and R. Leland, "Computing beyond Moore's Law," *Computer,* vol. 48, pp. 14-23, 2015.

[9] D. E. Nikonov and I. A. Young, "Overview of beyond-CMOS devices and a uniform methodology for their benchmarking," *Proceedings of the IEEE,* vol. 101, pp. 2498-2533, 2013.

[10] P. W. Shor, "Polynomial-time algorithms for prime factorization and discrete logarithms on a quantum computer," *SIAM review,* vol. 41, pp. 303-332, 1999.

[11] S. Agarwal, T.-T. Quach, O. Parekh, A. H. Hsia, E. P. DeBenedictis, C. D. James*, et al.*, "Energy Scaling Advantages of Resistive Memory Crossbar Based Computation and Its Application to Sparse Coding," *Frontiers in neuroscience,* vol. 9, 2015.

[12] A. P. Alivisatos, M. Chun, G. M. Church, K. Deisseroth, J. P. Donoghue, R. J. Greenspan*, et al.*, "The brain activity map," *Science (New York, NY),* vol. 339, p. 1284, 2013.

[13] H. Markram, "The human brain project," *Scientific American,* vol. 306, pp. 50-55, 2012.

[14] C. Bargmann, W. Newsome, A. Anderson, E. Brown, K. Deisseroth, J. Donoghue*, et al.*, "BRAIN 2025: a scientific vision," *Brain Research Through Advancing Innovative Neurotechnologies (BRAIN) Working Group Report to the Advisory Committee to the Director, NIH. Available online at: http://www. nih. gov/science/brain/2025/(US National Institutes of Health, 2014),* 2014.

[15] I. H. Stevenson and K. P. Kording, "How advances in neural recording affect data analysis," *Nature neuroscience,* vol. 14, pp. 139-142, 2011.

[16] N. Kasthuri, K. J. Hayworth, D. R. Berger, R. L. Schalek, J. A. Conchello, S. Knowles-Barley*, et al.*, "Saturated reconstruction of a volume of neocortex," *Cell,* vol. 162, pp. 648-661, 2015.

[17] J. Freeman, "Open source tools for large-scale neuroscience," *Current opinion in neurobiology,* vol. 32, pp. 156-163, 2015.

[18] P. Gao and S. Ganguli, "On simplicity and complexity in the brave new world of large-scale neuroscience," *Current opinion in neurobiology,* vol. 32, pp. 148-155, 2015.

[19] K. E. Bouchard, J. B. Aimone, M. Chun, T. Dean, M. Denker, M. Diesmann*, et al.*, "High-Performance Computing in Neuroscience for Data-Driven Discovery, Integration, and Dissemination," *Neuron,* vol. 92, pp. 628-631, Nov 2 2016.

[20] Y. LeCun, Y. Bengio, and G. Hinton, "Deep learning," *Nature,* vol. 521, pp. 436-444, 2015.

[21] A. K. Churchland and L. Abbott, "Conceptual and technical advances define a key moment for theoretical neuroscience," *Nature neuroscience,* vol. 19, pp. 348-349, 2016.

[22] R. Yuste, "From the neuron doctrine to neural networks," *Nature Reviews Neuroscience,* vol. 16, pp. 487-497, 2015.

[23] T. J. Sejnowski, P. S. Churchland, and J. A. Movshon, "Putting big data to good use in neuroscience," *Nature neuroscience,* vol. 17, pp. 1440-1441, 2014.

[24] X. Jiang, S. Shen, C. R. Cadwell, P. Berens, F. Sinz, A. S. Ecker*, et al.*, "Principles of connectivity among morphologically defined cell types in adult neocortex," *Science,* vol. 350, p. aac9462, 2015.

[25] W. Severa, O. Parekh, K. D. Carlson, C. D. James, and J. B. Aimone, "Spiking network algorithms for scientific computing," in *Rebooting Computing (ICRC), IEEE International Conference on*, 2016, pp. 1-8.

[26] B. M. Lake, R. Salakhutdinov, and J. B. Tenenbaum, "Human-level concept learning through probabilistic program induction," *Science,* vol. 350, pp. 1332-1338, 2015.

[27] S. Ahmad and J. Hawkins, "Properties of sparse distributed representations and their application to hierarchical temporal memory," *arXiv preprint arXiv:1503.07469,* 2015.

[28] W. Lotter, G. Kreiman, and D. Cox, "Deep predictive coding networks for video prediction and unsupervised learning," *arXiv preprint arXiv:1605.08104,* 2016.

[29] J. M. Kebschull, P. G. da Silva, A. P. Reid, I. D. Peikon, D. F. Albeanu, and A. M. Zador, "High-Throughput mapping of single-neuron projections by sequencing of barcoded RNA," *Neuron,* vol. 91, pp. 975-987, 2016.

[30] W. Maass, T. Natschläger, and H. Markram, "Real-time computing without stable states: A new framework for neural computation based on perturbations," *Neural computation,* vol. 14, pp. 2531-2560, 2002.

[31] H. Jaeger, "The "echo state" approach to analysing and training recurrent neural networks-with an erratum note," *Bonn, Germany: German National Research Center for Information Technology GMD Technical Report,* vol. 148, p. 13, 2001.

[32] C. Eliasmith and C. H. Anderson, *Neural engineering: Computation, representation, and dynamics in neurobiological systems*: MIT press, 2004.

[33] V. Mnih, K. Kavukcuoglu, D. Silver, A. A. Rusu, J. Veness, M. G. Bellemare*, et al.*, "Human-level control through deep reinforcement learning," *Nature,* vol. 518, pp. 529-533, 2015.

[34] T. J. Draelos, N. E. Miner, C. C. Lamb, C. M. Vineyard, K. D. Carlson, C. D. James*, et al.*, "Neurogenesis Deep Learning," *arXiv preprint arXiv:1612.03770,* 2016.

[35] G. Indiveri, B. Linares-Barranco, T. J. Hamilton, A. van Schaik, R. Etienne-Cummings, T. Delbruck*, et al.*, "Neuromorphic Silicon Neuron Circuits," *Frontiers in Neuroscience,* vol. 5, p. 73, 2011.

[36] S. H. Jo, T. Chang, I. Ebong, B. B. Bhadviya, P. Mazumder, and W. Lu, "Nanoscale memristor device as synapse in neuromorphic systems," *Nano letters,* vol. 10, pp. 1297-1301, 2010.

[37] C. D. James, J. B. Aimone, N. E. Miner, C. M. Vineyard, F. H. Rothganger, K. D. Carlson*, et al.*, "A historical survey of algorithms and hardware architectures for neural-inspired and neuromorphic computing applications," *Biologically Inspired Cognitive Architectures,* 2017.

[38] J. Cepelewicz, "The US Government Launches a $100-Million "Apollo Project of the Brain"," *Scientific American,* 2016.

[39] D. Marr, "Simple Memory: A Theory for Archicortex," *Philosophical Transactions of the Royal Society of London. Series B, Biological Sciences,* pp. 23-81, 1971.

[40] M. J. Milford, G. F. Wyeth, and D. Prasser, "RatSLAM: a hippocampal model for simultaneous localization and mapping," in *Robotics and Automation, 2004. Proceedings. ICRA'04. 2004 IEEE International Conference on*, 2004, pp. 403-408.

[41] J. B. Aimone and J. P. Weick, "Perspectives for computational modeling of cell replacement for neurological disorders," *Frontiers in computational neuroscience,* vol. 7, p. 150, 2013.

[42] P. A. Merolla, J. V. Arthur, R. Alvarez-Icaza, A. S. Cassidy, J. Sawada, F. Akopyan*, et al.*, "A million spiking-neuron integrated circuit with a scalable communication network and interface," *Science,* vol. 345, pp. 668-673, 2014.

[43] M. M. Khan, D. R. Lester, L. A. Plana, A. Rast, X. Jin, E. Painkras*, et al.*, "SpiNNaker: mapping neural networks onto a massively-parallel chip multiprocessor," in *Neural Networks, 2008. IJCNN 2008.(IEEE World Congress on Computational Intelligence). IEEE International Joint Conference on*, 2008, pp. 2849-2856.

[44] J. Schemmel, D. Brüderle, A. Grübl, M. Hock, K. Meier, and S. Millner, "A wafer-scale neuromorphic hardware system for large-scale neural modeling," in *Proceedings of 2010 IEEE International Symposium on Circuits and Systems*, 2010, pp. 1947-1950.